\documentclass[letterpaper, 10 pt, conference]{ieeeconf}  

\usepackage{graphicx}
\usepackage{amsmath}
\usepackage{booktabs}
\usepackage{multirow}
\usepackage{arydshln}
\usepackage{dblfloatfix}
\usepackage[style=numeric-comp,
            sorting=none,
            backend=bibtex]{biblatex}
\IEEEoverridecommandlockouts                              

\title{\LARGE \bf
GEM: Glare or Gloom, I Can Still See You -- \\
End-to-end Multi-modal Object Detection
}

\author{Osama Mazhar$^{1}$, Robert Babu\v{s}ka$^{1, 2}$ and Jens Kober$^{1}$ 
\thanks{The research presented in this article was carried out as part of the OpenDR project, which has received funding from the European Union’s Horizon 2020 research and innovation programme under Grant Agreement No. 871449.}
\thanks{$^{1}$Cognitive Robotics Department, Delft University of Technology, Delft, The Netherlands
        {\tt\small \{O.Mazhar, R.Babuska, J.Kober\}@tudelft.nl}}%
\thanks{$^{2}$Czech Institute of Informatics, Robotics, and Cybernetics, Czech Technical University in Prague, Czech Republic}%
}

\begin{document}

\maketitle
\thispagestyle{empty}
\pagestyle{empty}

\begin{abstract}

Deep neural networks designed for vision tasks are often prone to failure when they encounter environmental conditions not covered by the training data.
Single-modal strategies are insufficient when the sensor fails to acquire information due to malfunction or its design limitations.
Multi-sensor configurations are known to provide redundancy, increase reliability, and are crucial in achieving robustness against asymmetric sensor failures.
To address the issue of changing lighting conditions and asymmetric sensor degradation in object detection, we develop a multi-modal 2D object detector, and propose deterministic and stochastic sensor-aware feature fusion strategies.
The proposed fusion mechanisms are driven by the estimated sensor measurement reliability values/weights.
Reliable object detection in harsh lighting conditions is essential for applications such as self-driving vehicles and human-robot interaction.
We also propose a new ``r-blended'' hybrid depth modality for RGB-D sensors.
Through extensive experimentation, we show that the proposed strategies outperform the existing state-of-the-art methods on the FLIR-Thermal dataset, and obtain promising results on the SUNRGB-D dataset.
We additionally record a new RGB-Infra indoor dataset, namely L515-Indoors, and demonstrate that the proposed object detection methodologies are highly effective for a variety of lighting conditions.
\end{abstract}

\section{INTRODUCTION}

Modern intelligent systems such as autonomous vehicles or assistive robots should have the ability to reliably detect objects in challenging real-world scenarios.
Object detection is one of the widely studied problems in computer vision.
It has been addressed lately by employing deep convolutional neural networks where the state-of-the-art methods have achieved fairly accurate detection performances on the existing datasets~\cite{lin2017focal, liu2016ssd, redmon2018yolov3, ren2015faster}.
However, these vision models are fragile and
do not generalize across realistic unconstrained scenarios, such as changing lighting conditions or other environmental circumstances which were not covered by the training data \cite{yin2019fourier}.
The failure of the detection algorithms in such conditions could lead to potentially catastrophic results, as in the case of self-driving vehicles.

One way of addressing this problem is to employ a data-augmentation strategy \cite{hendrycks2020many}.
It refers to the technique of perturbing data without altering class labels, and it has been proven to greatly improve robustness and generalization performance~\cite{mazhar2021rsh}.
Nevertheless, this is insufficient for the cases where the sensor fails to acquire information due to malfunction or its technical limitations.
For example, the output of standard passive cameras degenerates with reduced ambient light, while thermal cameras or LiDARs are less affected by illumination changes.

Multi-sensor configurations are known to provide redundancy and often enhance the performance of the detection algorithms.
Moreover, efficient sensor fusion strategies minimize uncertainties, increase reliability, and are crucial in achieving robustness against asymmetric sensor failures~\cite{Bijelic2020a}.
Although, increasing the number of sensors might enhance the performance of detection algorithms, this comes with a considerable computational and energy cost.
This is not desirable in mobile robotic systems, which typically have constraints in terms of computational power and battery consumption.
In such cases, intelligent choice and combination of sensors are crucial.

\begin{figure}[t]
	\centering
	\includegraphics[width=0.99\linewidth, trim={0cm 0cm 0cm 0cm},clip]{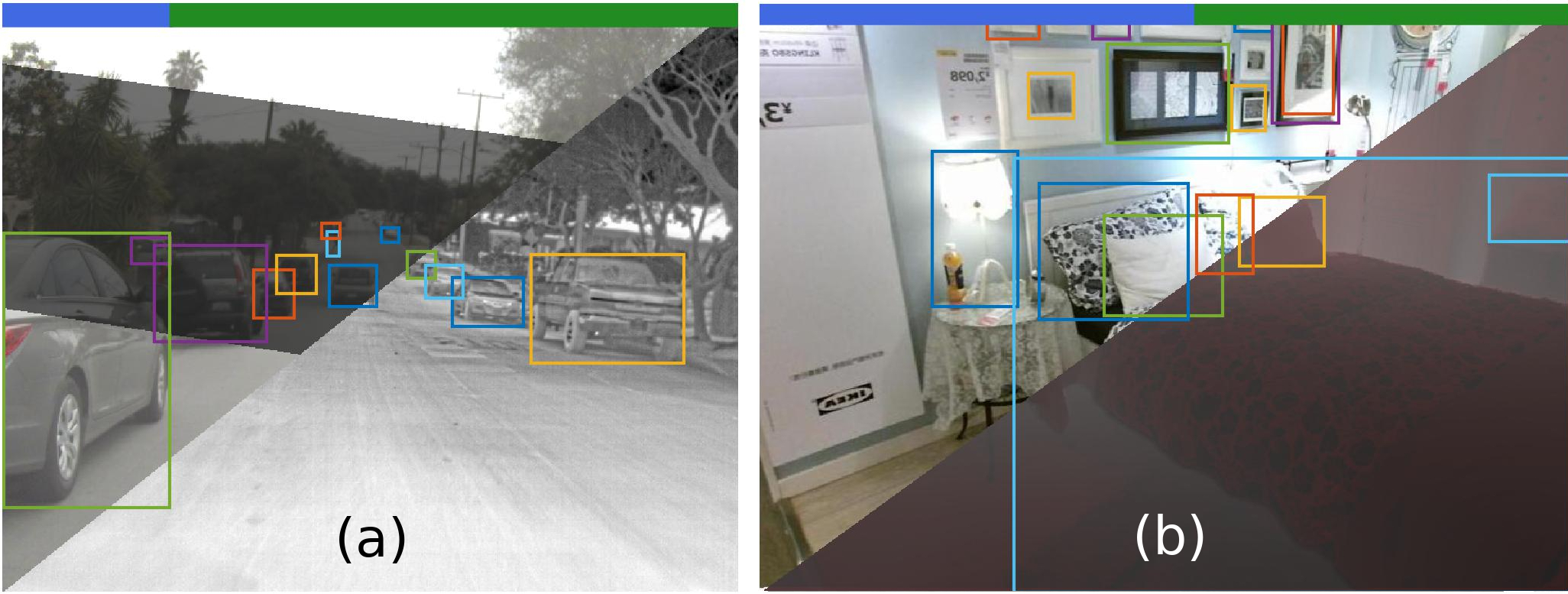}
	\caption{Output samples of the proposed multi-modal object detector.
	The blue/green bar at the top illustrates the contribution/reliability of each sensor modality in obtaining the final output. 
	Images from two modalities are merged diagonally only for illustration purposes. (a) Shows the results on the FLIR-Thermal dataset with RGB and thermal sensor modalities, (b) Shows the output on the SUNRGB-D dataset with RGB and our proposed ``r-blended'' hybrid depth modality.
	}
	\vspace{-0.5cm}
	\label{label1}
\end{figure}

Furthermore, multi-modal data fusion often requires an estimate of the sensor signal uncertainty to guarantee efficient fusion and reliable prediction without a priory knowledge of the sensor characteristics~\cite{martinez2017feature}.
The existing multi-modal object detection methods fuse the sensor data streams without explicitly modeling the measurement reliability.
This may have severe consequences when the data from an individual sensor degrades or is missing due to sheer sensor failure.

To address the above problems, we propose sensor-aware multi-modal fusion strategies for object detection in harsh lighting conditions, thus the title `` GEM: \underline{G}lare or gloom, I can still see you - \underline{E}nd-to-end \underline{M}ultimodal object detection''.
The output samples of GEM are shown in Figure~\ref{label1}.
Two fusion methods are proposed: deterministic weighted fusion and stochastic feature fusion.
In the deterministic weighted fusion, the measurement certainty of each sensor is estimated either by learning scalar weights or masks through separate neural networks.
The learned weights are then assigned to the feature maps extracted from the feature extractor backbones for each sensor modality.
The weighted feature maps can be fused either by averaging or concatenation.
Moreover, we can visualize and interpret the measurement certainty of each sensor in the execution phase, which provides deeper insights into the relative strengths of each data stream.
The stochastic feature fusion creates a one-hot encoding of the feature maps of each sensor, which can be assumed as a discrete switch that allows only the dominant/relevant features to pass.
The obtained selected features are then concatenated before they are passed to the object detection and classification head.
The proposed sensor-aware multi-modal object detector, referred to simply as GEM in the rest of the paper, is trained in an end-to-end fashion along with the fusion mechanism.

Most modern object detectors, including YOLO, Faster-RCNN and SSD employ many hand-crafted features such as anchor generation, rule-based assignment of classification and regression
targets as well as weights to each anchor, and non-maximum suppression postprocessing.
The overall performance of these methods often relies on careful tuning of the above-mentioned hyper-parameters.
Following their success in sequence/language modeling, transformers have lately emerged in vision applications, outperforming competitive baselines and demonstrating a strong potential in this field.
Therefore, we employ transformers in our work as in~\cite{Carion2020a}, which thanks to their powerful relational modeling capability eliminates the need of hand-crafted components in object detection.
Our main contributions in this paper are:
\begin{itemize}
    \item Evaluation of feature fusion in two configurations, i.e., deterministic weighted fusion and stochastic feature fusion for multi-modal object detection.
    \item Estimation of measurement reliability of each sensor as scalar or mask multipliers through separate neural networks for each modality to efficiently drive the deterministic weighted fusion.
    \item Use of transformers for multi-modal object detection to harness the efficacy of self-attention in sensor fusion.
    
\end{itemize}

\section{Related Work}
In this section, we first review deep learning-based object detection strategies, followed by a discussion on existing methods for multi-modal fusion methods in relevant tasks.

\subsection{Deep learning-based Object Detection}
Detailed literature surveys for deep learning-based object detectors have been published in \cite{Jiao2019a, Liu2020a}.
Here we briefly discuss some of the well-known object detection strategies.
Typically, object detectors can be classified into two types, namely two-stage and singe-stage object detectors.

\subsubsection{Two-stage object detection}

Two-stage object detectors exploit a region proposal network (RPN) in their first stage.
RPN ranks region boxes, alias \emph{anchors}, and proposes the ones that most likely contain objects as candidate boxes.
The features are extracted by region-of-interest pooling (RoIPool) operation from each candidate box in the second stage.
These features are then utilized for bounding-box regression and classification tasks.

\subsubsection{Single-stage object detection}

Single-stage detectors propose predicted boxes from input images in one forward pass directly, without the region proposal step.
Thus, this type of object detectors are time efficient and can be utilized for real-time operations.
Lately, an end-to-end object detection strategy has been proposed in \cite{Carion2020a} that eliminates the need for hand-crafted components like anchor boxes and non-maximum suppression.
The authors employ transformers in an encoder-decoder fashion, which have been extremely successful and become a de facto standard for natural language processing tasks.
The transformer implicitly performed region proposals
instead of using an R-CNN.
The multi-head attention module in transformers jointly attended to different semantic regions of an image/feature maps and linearly aggregates the outputs through learnable weights.
The learned attention maps can be visualized without requiring dedicated methods, as in the case of convolutional neural networks.
The inherent non-sequential architecture of transformers allows parallelization of models.
Thus, we opted to build upon the methodology of \cite{Carion2020a} for our multi-modal object detector for harsh lighting conditions.

\subsection{Sensor Fusion}







Sensor fusion strategies can be roughly divided into three types according to the level of abstraction where fusion is performed or in which order transformations are applied compared to feature combinations, namely low-level, mid-level, and high-level fusion \cite{Garcia2017a}.
In low-level or early fusion, raw information from each sensor is fused at pixel level, e.g., disparity maps in stereovision cameras \cite{Weichselbaum2013a}.
In mid-level fusion, a set of features is extracted for each modality in a pre-processing stage, while multiple approaches \cite{Park2016a} are exploited to fuse the extracted features.
Late-fusion often employs a combination of two fusing methods, e.g., convolution of stacked feature maps followed by several fully connected layers with dropout regularization \cite{Borghi2017a}.
In high-level fusion or ensemble learning methods, predictions are obtained individually for each modality and the learnt scores or hypotheses are subsequently combined via strategies such as weighted majority votes \cite{Nguyen2018a}.
Deep fusion or cross fusion \cite{Caltagirone2019a} is another type of fusion strategy which repeatedly combines inputs, then transforms them individually.
In each repetition, the transformation learns different features.
For example in
\cite{Bijelic2020a}, features from the layers of VGG network are  exchanged  among  all  modalities  driven  by sensor  entropy  after  each  pooling  operation.

\subsection{Multi-sensor Object Detection}

Most of the efforts on multi-modal object detection in the literature are focused on pedestrian or vehicle detection in the automotive context.
Sensor fusion strategies are typically proposed for camera-LiDAR, camera-radar, and camera-radar-LiDAR setups.
Here, we briefly go through the relevant state-of-the-art methods.
\begin{figure}[t]
	\centering
	\includegraphics[width=0.99\linewidth, trim={0cm 0cm 0cm 0cm},clip]{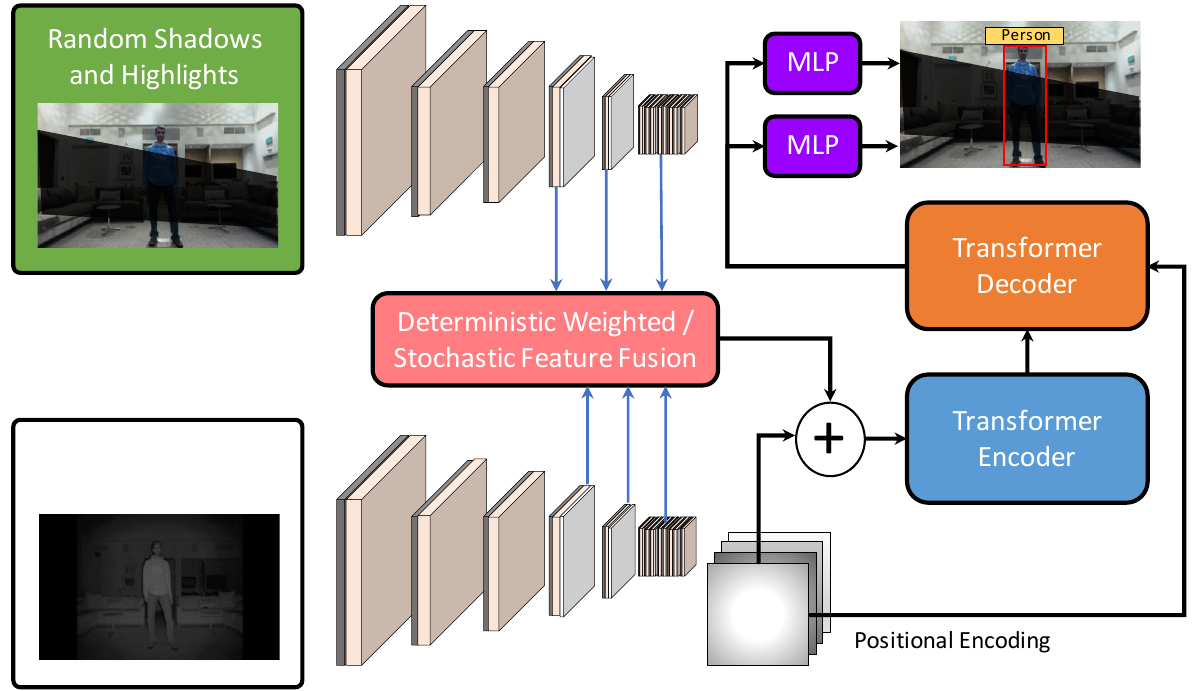}
	\caption{Our proposed pipeline for a multi-modal object detector with transformers.
	The features from each backbone are fused and passed to the transformer encoder-decoder network.
	The decoder output is subsequently exploited by Multilayer Perceptrons (MLPs) for bounding box regression and object classification.}
	\label{TUD:overallmmod}
	\vspace{-0.5cm}
\end{figure}

The authors in \cite{Bijelic2020a} proposed an entropy-steered multi-modal deep fusion architecture for adverse weather conditions.
The sensor modalities exploited in their method include RGB camera, gated camera (NIR band), LiDAR, and radar.
Instead of employing BeV projection or point cloud representation for LiDAR, the authors encoded depth, height, and pulse intensity on an image plane.
Moreover, the radar output was also projected onto an image plane parallel to the image horizontal dimension.
Considering the radar output invariant along the vertical image axis, the scan was replicated across the horizontal image axis.
They utilized a modified VGG architecture for feature extraction, while features were exchanged among all modalities driven by sensor entropy after each pooling operation.
Fused feature maps from the last 6 layers of the feature extractors were passed to the SSD object detection head.
In \cite{Devaguptapu2019a}, the authors proposed a pseudo multi-modal object detector from thermal IR images in a Faster-RCNN setting.
The features from ResNet-50 backbones for the two modalities are concatenated and a $1\times1$ convolution is applied to the concatenated features before they are passed to the rest of Faster-RCNN network.
They exploited I2I translation networks, namely CycleGAN~\cite{isola2017image} and UNIT~\cite{liu2017unsupervised} to transform thermal images from the FLIR Thermal~\cite{Flir} and KAIST~\cite{hwang2015multispectral} datasets to the RGB domain, thus the names MM-CG and MM-UNIT.
Here, we also discuss some fusion strategies which were originally proposed for applications other than object detection but are relevant to our work.
In \cite{Khan2019a}, the authors proposed a sensor fusion methodology for RGB and depth images to steer a self-driving vehicle.
The latent semantic vector from an encoder-decoder segmentation network trained on RGB images was fused with the depth features.
The fusion architecture proposed by \cite{Khan2019a} is similar to the gating mechanism driven by the learned scalar weights presented in \cite{Patel2017a}.
The method proposed in \cite{Chen2019a} is closest to our work.
The authors proposed two sensor fusion strategies for Visual-Inertial Odometry (VIO), namely soft fusion and hard fusion.
In soft fusion,
they learned soft masks which were subsequently assigned to each element in the feature vector.
Hard fusion employed a variant of the Gumbel-max trick, which is often used to sample discrete data from categorical distributions.
Learning masks equal to the size of feature vectors might introduce computational overhead.
Therefore, we learn dynamic scalar weights for each sensor modality, which adapt to the environmental/lighting conditions.
These scalar weights represent the reliability or relevance of the sensor signals.
Moreover, we also learn single-channel masks with a spatial size equal to that of the feature maps obtained from the feature extractor backbone.
Nevertheless, we also implement the Gumbel-Softmax trick for comparison, as a stochastic feature fusion for multi-modal object detectors.





\section{Sensor-aware Multi-modal Fusion}

In this work, we propose a new method for sensor-aware feature selection and multi-modal fusion for object detection.
We actually evaluate feature fusion in two configurations, i.e., deterministic weighted fusion with scalar and mask multipliers, and stochastic feature fusion driven by the Gumbel-Softmax trick that enables sampling from a discrete distribution.
The overall pipeline of the proposed multi-modal object detector is illustrated in Figure~\ref{TUD:overallmmod}.
The proposed methodologies are trained and evaluated on datasets with RGB and thermal or depth images.
However, it can be extended to include data from other sensors like LiDAR or radar, either by projecting the sensor output onto an image plane as proposed in \cite{Bijelic2020a} or by employing sensor-specific feature extractors such as~\cite{Charles2017a}.




\subsection{Deterministic Weighted Fusion}

The proposed deterministic weighted fusion scheme is conditioned on the measurement certainty of each sensor.
These values are obtained either by learning scalar weights or masks through separate neural networks.
Subsequently, the weights are assigned to the feature maps extracted from the backbones as (scalar or mask) multipliers for each sensor modality.
Given the output of the backbone feature extractors $\mathbf{s}$ for a single modality, the neural network $f$ optimizes parameters $\boldsymbol{\theta}$ to obtain measurement certainty $w$ of the corresponding sensor as described as follows:
\begin{gather}
w = f(\mathbf{s}, \boldsymbol{\theta}) \times \frac{1}{\text{rows}\times \text{cols} \times k}\sum_{l=1}^\text{rows}\sum_{m=1}^\text{cols}\sum_{n=1}^{k} \mathbf{s}(l,m,n)
\label{eqn:weight_estimation}
\end{gather}
where $k$ is the selected number of channels.
The network $f$ learns the parameters $\boldsymbol{\theta}$ in an end-to-end fashion.
In the case of sensor degradation, the output of the neurons in the early layers of the corresponding backbone will remain close to zero.
Thus, we multiply the output of the network $f$ by the mean of first $k$ feature maps, $16$ in our case, from $\mathbf{s}$ in a feed-forward setting to obtain $w$.
This allows $f$ to dynamically condition its output to changing lighting/sensor degradation scenarios, which subsequently guides the transformers to focus on the dominant sensor modality for object detection.
Furthermore, the multiplication of the raw output of $f$ with the mean of the selected feature maps is performed without gradient calculation to prevent the distortion of the feature maps in the back-propagation phase.

The weighted feature maps are fused by either taking an average of the two feature sets, or by concatenating them.
The fused features are then passed to the transformer for object detection and localization.
Our scalar fusion functions $g_\text{sa}$ (averaging) and $g_\text{sc}$ (concatenating) are represented as:
\begin{gather}
    g_\text{sa}(\mathbf{s_\text{RGB}}, \mathbf{s_\text{IR}}) = \phi(w_\text{RGB} \odot \mathbf{s_\text{RGB}}, w_\text{IR} \odot \mathbf{s_\text{IR}}) \nonumber \\
    g_\text{sc}(\mathbf{s_\text{RGB}}, \mathbf{s_\text{IR}}) = [w_\text{RGB} \odot \mathbf{s_\text{RGB}}; w_\text{IR} \odot \mathbf{s_\text{IR}}]
    \label{eqn:scalarfusion}
\end{gather}
where $\phi$ denotes the mean operation, $\mathbf{s_\text{RGB}}$ and $\mathbf{s_\text{IR}}$ are feature maps obtained from the backbone feature extractor for RGB and thermal/IR imagers respectively, while $w_\text{RGB}$ and $w_\text{IR}$ are the sensor measurement certainty weights obtained through Equation~\eqref{eqn:weight_estimation}.
Similar to the scalar fusion method, feature selection is also modelled by learning masks for each modality, in this case $\mathbf{m_{\text{RGB}}}$ and $\mathbf{m_{\text{IR}}}$/$\mathbf{m_{\text{depth}}}$, with a spatial size equal to that of the features maps.
The fusion scheme with mask multipliers is represented as:
\begin{gather}
    g_\text{ma}(\mathbf{s_\text{RGB}}, \mathbf{s_\text{IR}}) = \phi(\mathbf{m_\text{RGB}} \odot \mathbf{s_\text{RGB}}, \mathbf{m_\text{IR}} \odot \mathbf{s_\text{IR}}) \nonumber \\
    g_\text{mc}(\mathbf{s_\text{RGB}}, \mathbf{s_\text{IR}}) = [\mathbf{m_\text{RGB}} \odot \mathbf{s_\text{RGB}}; \mathbf{m_\text{IR}} \odot \mathbf{s_\text{IR}}]
    \label{eqn:maskfusion}
\end{gather}

\subsection{Stochastic Feature Fusion}
In addition to the weighted fusion schemes, we exploit a variant of the Gumbel-max trick to learn a one-hot encoding that either propagates or blocks each component of the feature maps for intelligent fusion.
The Gumbel-max resampling strategy allows to draw discrete samples from a categorical distribution during the forward pass through a neural network.
It exploits the reparametrization trick to separate out the deterministic and stochastic parts of the sampling process.
However, it adds Gumbel noise instead of that from a normal distribution, which is actually used to model the distribution of the maximums for samples taken from other distributions.
Gumbel-max then employs the $\arg\max$ function to find the class that has the maximum value for each sample.

Considering $\alpha$ be the $n$-dimensional probability variable conditioned for every row on each channel of the feature volume such that $\alpha=[\pi_1,\ldots,\pi_n]$, representing the probability of each feature at location $n$, the Gumbel-max trick can be represented by the following equation:
\begin{equation} \label{TUD:gumbel_argmax}
Q = \arg\max_{i}(\log\pi_i + G_i)
\end{equation}
where, $Q$ is a categorical variable with class probabilities $\pi_1, \pi_2, \dots, \pi_n$
and $\{G_i\}_{i\leq{n}}$ is an i.i.d.\ sequence of standard Gumbel random variables which is given by:
\begin{equation} \label{TUD:Gumbel_dist}
G = -\log(-\log(U)), \quad U\sim \operatorname{Uniform}[0, 1] 
\end{equation}

The use of $\arg\max$ makes the Gumbel-max trick \emph{non-}differentiable.
However, it can be replaced by \emph{Softmax} with a temperature factor $\tau$, thus making it a fully-differentiable resampling method \cite{Jang2016a}.
Softmax with temperature parameter $\tau$ can be represented as
\begin{equation} \label{TUD:softmax}
f_\tau(x)_i = \frac{\text{exp}(x_i/\tau)}{\Sigma_{j=1}^n\text{exp}(x_j/\tau)} 
\end{equation}
where $\tau$ determines how closely the Gumbel-Softmax distribution matches the categorical distribution.
With low temperatures, e.g., $\tau=0.1$ to $\tau=0.5$, the expected value of a Gumbel-Softmax random variable approaches the expected value of a categorical random variable \cite{Jang2016a}.
The Gumbel-Softmax resampling function can therefore be written as
\begin{equation} \label{TUD:Gumbel-softmax}
Q^{\tau}_i=f_{\tau}(\log\pi_i+G_i)=\frac{\text{exp}((\log\pi_i+G_i)/\tau)}{\Sigma_{j=1}^{n}\text{exp}((\log\pi_j+G_j)/\tau)}
\end{equation}
with $i=1,\ldots,n$.

\begin{figure}[t]
	\centering
	\includegraphics[width=0.7\linewidth, trim={0.5cm 0.6cm 0.2cm 0.6cm},clip]{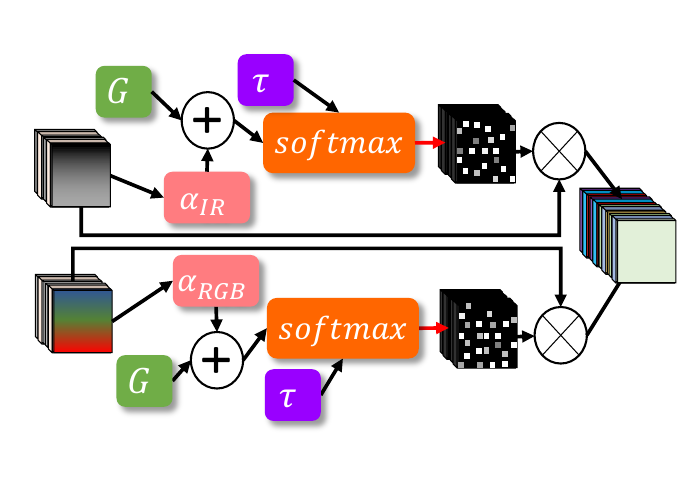}
	\caption{Illustration of our stochastic feature fusion strategy that employs the Gumbel-Softmax sampling trick.}
	\vspace{-0.5cm}
	\label{TUD:gumbel_softmax}
\end{figure}

We set $\tau=1$ and obtain feature volume approximate one-hot categorical encodings for each modality $\mathbf{e_{RGB}}$ and $\mathbf{e_{IR}}$.
Then a Hadamard product is taken between the encodings and the feature volumes and the resultants are subsequently concatenated and passed on to the bounding box regressor and classification head.
We illustrate our selective fusion process developed for multi-modal object detector in Figure~\ref{TUD:gumbel_softmax}, while the selective fusion function $g_\text{sf}$ is given as follows
\begin{equation} \label{TUD:selective_fusion}
g_\text{sf}(\mathbf{s_\text{RGB}}, \mathbf{s_\text{IR}}) = [\mathbf{e_\text{RGB}} \odot \mathbf{s_\text{RGB}}; \mathbf{e_\text{IR}} \odot \mathbf{s_\text{IR}}].
\end{equation}

\vspace{-0.4cm}
\section{Experiments}
\subsection{Datasets}
Three datasets are utilized in the training and evaluation of GEM, including the FLIR Thermal, SUNRGB-D \cite{Shuran2015a} and a new L515-Indoor dataset that we recorded for this research.
The FLIR Thermal dataset provides 8,862 training and 1,366 test samples of thermal and RGB images recorded in the streets and highways in Santa Barbara, California, USA.
Only the thermal images in the dataset are annotated with four classes, i.e., \emph{People}, \emph{Bicycle}, \emph{Car} and \emph{Dog}.
The given RGB images in the dataset are neither annotated nor aligned with their thermal counterparts, while the camera matrices are also not provided.
%
Thus, to utilize this dataset in a multi-modal setting, the given RGB images must be annotated or aligned with their corresponding thermal images.
One way to address this problem is to create artificial RGB images from input thermal images through GANs or similar neural networks as performed in \cite{Devaguptapu2019a}.
However, we opted to employ the concept of homography by manually selecting matching features in multiple RGB and thermal images.
The selected feature points are then employed to estimate a transformation matrix between the two camera modalities.
RGB images are subsequently transformed with the estimated homography matrix such that they approximately align with their thermal equivalents.
The \emph{Dog} class constitutes only $0.29\%$ of all the annotations in the aligned FLIR-Thermal dataset, thus it is not included in our experiments.

The SUNRGB-D dataset contains 10,335 RGB-D images taken by Kinect v1, Kinect v2, Intel RealSense, and Asus Xtion cameras. 
The annotations provided consist of 146,617 2D polygons and 64,595 3D bounding boxes, while 2D bounding boxes are obtained by projecting the coordinates of 3D bounding boxes onto the image plane.
Although the dataset contains labels for approximately 800 objects, we evaluate our method on the selected 19 objects similar to \cite{Shuran2015a}.
We first divide the dataset into three subsets such that the training set consists of 4,255 images, the validation set has 5,050 images, while the test set contains 1,059 images.

\begin{figure}[!t]
	\centering
	\includegraphics[width=0.99\linewidth, trim={0cm 0.2cm 0cm 0.4cm},clip]{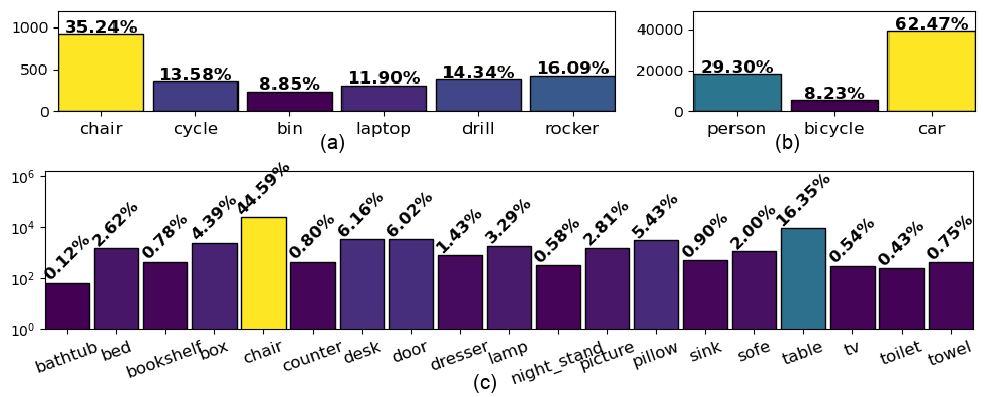}
	\vspace{-0.7cm}
	\caption{Class distributions of the datasets (a) L515-Indoor (b) FLIR-Thermal (c) SUNRGB-D. The number of annotations in (c) are presented in the logarithmic scale. }
	\label{data_distribution}
	\vspace{-0.6cm}
\end{figure}

The L515-Indoor dataset provides 482 training and 207 validation RGB and IR images recorded with Intel RealSense L515 camera with various ambient light conditions in an indoor scene.
It contains annotations of 1,819 2D bounding boxes of 6 object categories in total.
The IR images are aligned with their RGB counterparts through a homography matrix which is computed in a similar fashion as explained for the FLIR-Thermal dataset.
The population distributions of the datasets are illustrated in Figure~\ref{data_distribution}.

\subsection{Pre-processing Sensor Outputs}
For the FLIR-Thermal and L515-Indoor datasets, aligned RGB and thermal/IR images are fed into our feature extractor backbones without any pre-processing.
However, techniques that exploit datasets with depth images including \cite{Shuran2015a} often apply HHA encoding \cite{Gupta2014a} on the depth sensor modality for early feature extraction prior to being fed into the neural networks.
HHA is a geocentric embedding for depth images that encodes horizontal disparity, height above ground, and angle with gravity for each pixel.
In a multi-threading setup on a 12-Core Intel\textsuperscript{\textregistered} Core\textsuperscript{TM} i7-9750H CPU, HHA encoding of a batch of $32$ images takes approximately $119$ seconds, which is far from its application in real-time object detection or segmentation tasks.

To address this problem, assuming that we are working with RGB and depth modalities, we create a new hybrid image that introduces scene texture in a depth image. As the red light is scattered the least by air molecules, we blend the depth images and the red image channels through a blend weight $\alpha$.
Thus, we name our hybrid depth image as ``r-blended'' depth image.
\begin{gather}
\text{img}_{\text{r-blended}} = \alpha~ \text{img}_\text{depth} + (1 - \alpha)~\text{img}_{\text{red}}
\end{gather}
The value of $\alpha$ is set to $0.9$ for depth images while the weight value for the red channels becomes $0.1$.
This is to make sure that when the neural network is trained with ``r-blended depth'' image, it should focus on learning the depth features while information from the red channel only complements the raw depth map.
The idea to blend the red channel is also supported by the fact that CMOS cameras are often more sensitive to green and red light.
We first train our multi-modal object detector on RGB and HHA encoded depth images.
Later, we fine-tune the trained model by replacing HHA encoded images with r-blended depth images and achieve comparable results in terms of detection accuracy, while the fine-tuned model can indeed be used for real-time multi-modal object detection.

\begin{table*}[!b]
\vspace{-0.35cm}
\renewcommand\thetable{II} 
\centering
\caption{Performance evaluation on SUNRGB-D dataset.}
\label{tab:SUNRGB-D}
\begin{tabular}{c|c|c|c|c|c|c|c|c|c|c|c|c} 
\hline
\multirow{3}{*}{Models}                 & \multicolumn{11}{c|}{Tested without Random Shadows and Highlights (RSH)}                                                                                          & \multicolumn{1}{c}{w/ RSH}                                                                   \\ 
\cline{2-13}
                                        & \multicolumn{10}{c|}{AP@IoU=0.5}                                                        & \multirow{2}{*}{\begin{tabular}[c]{@{}c@{}}mAP@\\IoU=0.5 \end{tabular}} & \multicolumn{1}{c}{\multirow{2}{*}{\begin{tabular}[c]{@{}c@{}}mAP@\\IoU=0.5 \end{tabular}}}  \\ 
\cline{2-11}
                                        & bathtub & bed   & bookshelf & box   & chair & \dots & door  & dresser & lamp  & night stand &                                                                         & \multicolumn{1}{c}{}                                                                         \\ 
\hline\hline
RGB-only                                & 0.116   & 0.461 & 0.038     & 0.084 & 0.457 & \dots & 0.370 & 0.085   & 0.185 & 0.095       & 0.224                                                                   & 0.169                                                                                         \\
HHA-only                                & 0.355   & 0.409 & 0.002     & 0.020 & 0.413 & \dots & 0.113 & 0.024   & 0.199 & 0.057       & 0.165                                                                   & 0.093                                                                                         \\ 
\hdashline
conc-baseline                           & 0.333       & 0.440     & 0.002        & 0.068     & 0.456     & \dots & 0.367     & 0.056       & 0.195    & 0.041          & 0.211                                                                      & 0.155                                                                                             \\ 
avg-baseline                                  & 0.174   & 0.461 & 0.032          & 0.062          & 0.470          &  & 0.339          & 0.030          & 0.220          & 0.049          & 0.207                                                                  & 0.166                                                                                        \\ 
\hdashline
$g_\text{sc}$(raw-depth)                         & 0.404   & 0.411 & 0.008     & 0.073 & 0.487 & \dots & 0.360 & 0.051   & 0.225 & 0.044       & 0.226                                                                   & 0.209                                                                                            \\
$g_\text{sc}$(r-blended)  & 0.350   & 0.457 & 0.040     & 0.085 & \textbf{0.490} & \dots & \textbf{0.381} & \textbf{0.107}   & \textbf{0.226} & \textbf{0.108}       & \textbf{0.242}                                                                   & 0.230                                                                                         \\
\hdashline
$g_\text{sa}$(raw-depth) & 0.204   & 0.399 & 0.033          & \textbf{0.087} & 0.478          &  & 0.344          & 0.102          & 0.220          & 0.025          & 0.221                                                                  & 0.214                                                                                        \\
$g_\text{sa}$(r-blended) & 0.253   & 0.423 & \textbf{0.106} & 0.080          & 0.474          &  & 0.379          & 0.035          & 0.219          & 0.079          & 0.239                                                                  & \textbf{0.236}                                                                               \\
\hline
\end{tabular}
\end{table*}

\begin{table}[!t]
\renewcommand\thetable{I} 
\centering
\caption{Performance evaluation on FLIR-THERMAL dataset.}
\label{tab:FLIR}
\begin{tabular}{c|c|c|c|c|c} 
\hline
\multirow{3}{*}{\begin{tabular}[c]{@{}c@{}}Model\\w/ RSH\end{tabular}} & \multicolumn{4}{c|}{w/o Random Shadows and Highlights}                                                                    & \multicolumn{1}{c}{w/ RSH}                                                                  \\ 
\cline{2-6}
                                                                       & \multicolumn{3}{c|}{AP@IoU=0.5}                  & \multirow{2}{*}{\begin{tabular}[c]{@{}c@{}}mAP@\\IoU=0.5\end{tabular}} & \multicolumn{1}{c}{\multirow{2}{*}{\begin{tabular}[c]{@{}c@{}}mAP@\\IoU=0.5\end{tabular}}}  \\ 
\cline{2-4}
                                                                       & Person         & Bicycle        & Car            &                                                                        & \multicolumn{1}{c}{}                                                                        \\ 
\hline\hline
FLIR Baseline                                                          & 0.794          & 0.580          & 0.856          & 0.743                                                                  & -                                                                                            \\ 
\hdashline
rgb-only                                                               & 0.383          & 0.168          & 0.638          & 0.395                                                                  & 0.376                                                                                        \\
thermal-only                                                           & 0.683          & 0.499          & 0.783          & 0.655                                                                  & 0.316                                                                                        \\ 
\hdashline
MM-UNIT \cite{Devaguptapu2019a}                                                                & 0.644          & 0.494          & 0.707          & 0.615                                                                  & -                                                                                            \\
MM-CG \cite{Devaguptapu2019a}                                                                 & 0.633          & 0.502          & 0.706          & 0.614                                                                  & -                                                                                            \\ 
\hdashline
SSD-BL                                                       & 0.450              & 0.341              & 0.719              & 0.503                                                                      & 0.478                                                                                        \\
SSD-WA                                                       & 0.526          & 0.314          & 0.718          & 0.519                                                                  & 0.516                                                                                        \\ 
\hdashline
avg-baseline                                                           & 0.801          & 0.562          & 0.879          & 0.747                                                                  & 0.731                                                                                        \\
conc-baseline                                                          & 0.533          & 0.417          & 0.675          & 0.541                                                                  & 0.492                                                                                        \\ 
\hdashline
$g_\text{sa}$                                                          & \textbf{0.828} & 0.593          & \textbf{0.891} & \textbf{0.770}                                                         & \textbf{0.769}                                                                               \\
$g_\text{sc}$                                                          & 0.809          & \textbf{0.637} & 0.862          & 0.769                                                                  & 0.764                                                                                        \\
$g_\text{ma}$                                                          & 0.803          & 0.575          & 0.862          & 0.746                                                                  & 0.744                                                                                        \\
$g_\text{mc}$                                                          & 0.800          & 0.611          & 0.857          & 0.755                                                                  & 0.755                                                                                        \\
$g_\text{sf}$                                                          & 0.790          & 0.584          & 0.874          & 0.749                                                                  & 0.756                                                                                        \\ 
\hdashline
$g_\text{sc}$ m-net                                                 & 0.696          & 0.472          & 0.823          & 0.663                                                                  & 0.664                                                                                        \\
\hline
\end{tabular}
\vspace{-0.7cm}
\end{table}
\vspace{-0.1cm}
\subsection{Training}
GEM is trained with scalar fusion and mask fusion methods, i.e., $g_\text{sa}$, $g_\text{sc}$, $g_\text{ma}$ and $g_\text{mc}$ for deterministic weighted fusion driven by Equations~\eqref{eqn:scalarfusion} and~\eqref{eqn:maskfusion}, while it is also trained with $g_\text{sf}$ for stochastic feature fusion.
The backbone feature extractors for both sensor streams and the transformer block are pre-trained on MSCOCO dataset on RGB images as in~\cite{Carion2020a}.
For the FLIR thermal dataset, each model is trained on a cluster with 2 GPUs for 100 epochs while the models for the SUNRGB-D dataset are trained with 4 GPUs for 300 epochs.
Similarly, the models for L515-Indoor are trained on a cluster with 2 GPUs for 300 epochs with a batch size of 1. 
The batch size for the FLIR-Thermal and SUNRGBD datasets is set to 2, while the learning rate for the feature extractor backbones, fusion networks, and transformer block is set to $8 \times 10^{-6}$ for all datasets.
We employ ResNet-50 as the feature extractor, while we also train $g_\text{sc}$ on MobileNet v2~\cite{howard2017mobilenets} for the FLIR thermal dataset.
To guide the fusion process and mimic harsh lighting conditions for the RGB sensor, we also employ Random Shadows and Highlights (RSH) data augmentation as proposed in \cite{mazhar2021rsh}.
RSH develops immunity against lighting perturbations in the convolutional neural networks, which is desirable for real world applications.
We additionally implement SSD512 object detector with VGG16 backbones in a multi-modal setting in two configurations, i.e., a simple averaging fusion as the baseline method (SSD-BL) and a weighted averaging fusion scheme (SSD-WA) similar to $g_{\text{sa}}$.
The anchor/default boxes are configured for both SSD-BL and SSD-WA in a fashion similar to that for the MS-COCO dataset.
These models are trained for 800 epochs in a single GPU setup with a batch size 1 and a learning rate $1 \times 10^{-4}$ which decays with a decaying factor 0.2 after the first 520,000 iterations.

\subsection{Evaluation}
\subsubsection*{FLIR-Thermal} performance evaluation of the proposed networks on the FLIR-Thermal dataset is shown in Table~\ref{tab:FLIR}.
We show Average Precision (AP) values at Intersection over Union (IoU) of $0.5$ for each dominant class, while the mean Average Precision (mAP) is also estimated with and without lighting perturbations. 
These lighting corruptions are introduced by creating Random Shadows and Highlights (RSH) on the test RGB images.
The evaluation with lighting perturbation is performed for 10 trials in all experiments, while the average of the obtained mAP is shown in the table.
The results are compared with the single modality object detector, the multi-modal baseline fusion networks, and the existing state-of-the-art methods on this dataset.
In the baselines, the features from the backbones are fused in two configurations: averaged and concatenated, without any weighing or re-sampling mechanism.
Additionally, we compare the performance of SSD-BL and SSD-WA on the FLIR-Thermal dataset.
It is clear from the evaluation results, that our proposed methodologies, i.e., $g_\text{sa}$, $g_\text{sc}$, $g_\text{ma}$, $g_\text{mc}$, and $g_\text{sf}$, outperform the previously reported results on this dataset.
Our methods also demonstrate robustness against lighting perturbation, while a significant performance drop of single modality and baseline methods can be seen when tested with RSH.
The ``avg-baseline'' obtained comparable results, but as it is only a blind fusion, hence no sensor contribution or reliability measure can be obtained with this methodology.
Additionally, its performance can significantly drop in the case of asymmetric sensor failure.
This can partially be observed in Table~\ref{tab:FLIR} where the baselines are tested with RSH perturbations.
Concerning the evaluation of multi-modal SSD on the FLIR-Thermal dataset, SSD-WA certainly improves the results compared to SSD-BL, specifically in terms of robustness against lighting perturbations introduced by RSH.
The overall performance of SSD-based detectors turned out to be inferior to that of our transformer-based multi-modal object detection methods.

Among our proposed fusion methods, $g_\text{sa}$ obtained the best overall performance on the FLIR-Thermal dataset.
Scalar multiplication amplifies the information in the feature maps by retaining the learned structure.
Nevertheless, mask multiplication may amplify a certain spatial portion of the feature maps in some channels, but it could also potentially distort the learned information depth-wise.
Concatenation might be useful when the feature spaces of the utilized sensor modalities differ, e.g., image versus point cloud.
However, in our case of image modalities, the averaging features $g_\text{sa}$ performed better than concatenation $g_\text{sc}$. 
Similarly, switching off the features with selective fusion $g_\text{sf}$ has affected the performance of the model adversely.
We plan to explore this method further in our future research, especially in the cases when information from the sensor modalities of dissimilar domains are fused, e.g., camera versus LiDAR/radar.

\begin{figure*}[!t]
	\centering
	\includegraphics[width=0.99\linewidth, trim={0cm 0cm 0cm 0cm},clip]{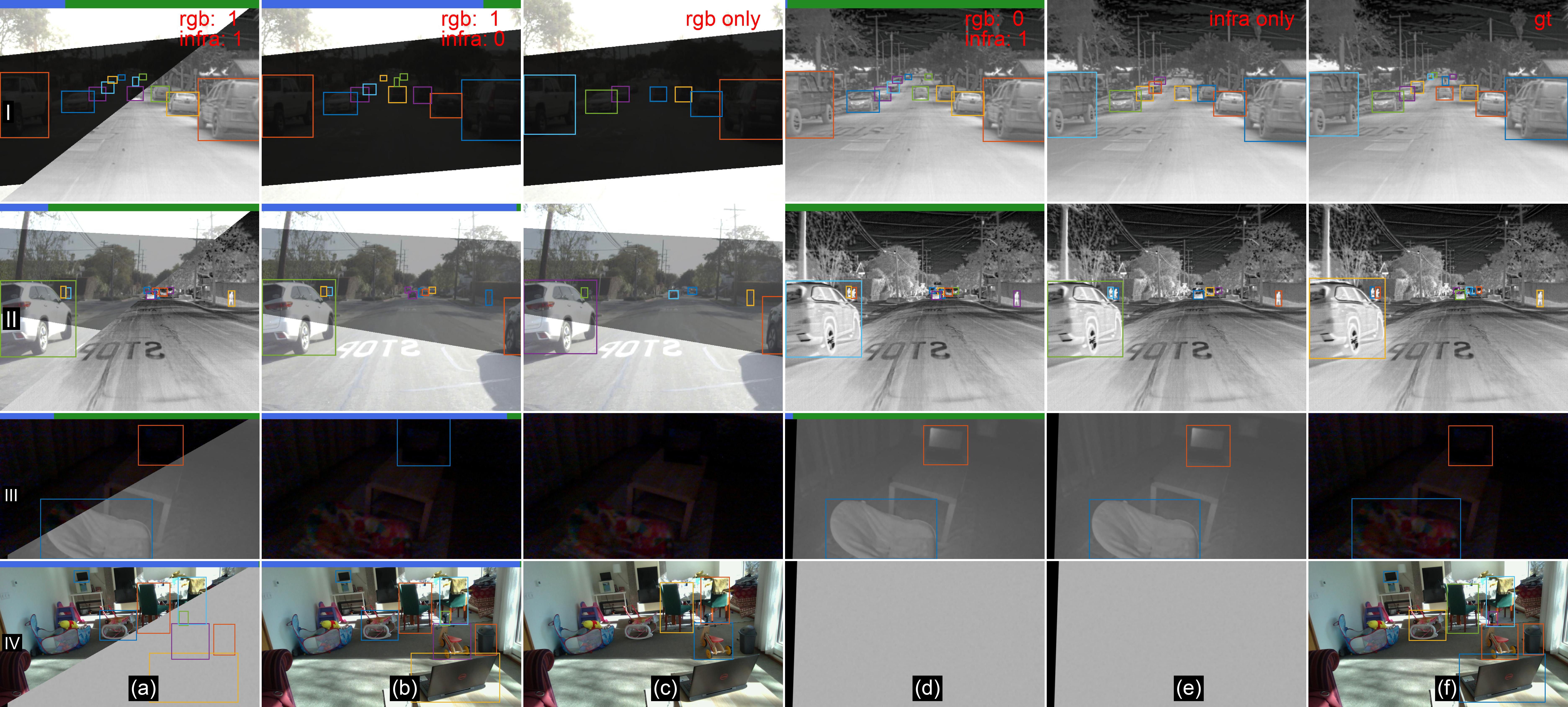}
	\vspace{-0.2cm}
	\caption{Qualitative analysis of our multi-modal object detector, $g_\text{sa}$ in this case.
	Columns (a), (b) and (d) are the outputs of $g_\text{sa}$ in various asymmetric sensor failure conditions imitated artificially, which are mentioned on the upper-right corner of each image in row I.
	The top blue/green bar represents the contribution of each sensor modality in obtaining the final results (RGB: blue and Thermal/Infra: green).
	(c) and (e) are the outputs from single modal baselines.
	(f) is the ground-truth.
	Rows I and II are from FLIR-Thermal dataset while III and IV are from L515-Indoor dataset.
	Row IV represents a true sensor failure case when IR camera gets saturated due to sun-light even in indoors.
	}
	\label{Fig:qualitative_FLIR_L515}
	\vspace{-0.5cm}
\end{figure*}

\subsubsection*{SUNRGB-D} The evaluation results on SUNRGB-D dataset are shown in Table~\ref{tab:SUNRGB-D}.
We not only present a comparison of single vs.\ multi-modal settings on the selected 19 categories of the SUNRGB-D dataset, but also between raw vs.\ processed depth images.
The table only shows the results for eight categories due to limited space.
Two single modality networks are trained, one with RGB images and the other with HHA-encoded depth images.
We also evaluate the performance of ``conc-baseline'' and ``avg-baseline'' with RGB and HHA-encoded depth modalities.
Motivated by the performance of $g_\text{sa}$ and $g_\text{sc}$ on the FLIR-Thermal dataset, we chose to evaluate their performance on SUNRGB-D dataset exclusively.
Since HHA-encoding introduces a significant computational burden inhibiting the possibility of real-time object detection, we first train $g_\text{sa}$ and $g_\text{sc}$ with on RGB and HHA-encoded depth images, later we fine-tune these models on raw-depth images as well as on our ``r-blended'' hybrid depth images.
It is evident in Table~\ref{tab:SUNRGB-D} that both $g_\text{sa}$ and $g_\text{sc}$ obtain promising results on this dataset with RGB and ``r-blended'' depth images.
Further analysing the results of Table~\ref{tab:FLIR}, we observe that the comparative performance of the models on the {\em Bicycle} class is not stable.
Looking at the distribution of the datasets in Figure~\ref{data_distribution}, we realize that the {\em Bicycle} class only constitutes 8.23\% of the dataset.
This indicates its comparative inconsistent performance on various models.
However, analysing the results in Table~\ref{tab:SUNRGB-D}, we realize this performance instability might also be related to the object size.
The proposed networks are able to distinguish large sized objects even if their contribution in the dataset is relatively small e.g., {\em Baththub} and {\em Bed} classes.
This problem can be traced back to \cite{Carion2020a} which itself struggles to perform equally on detecting small sized objects.

\subsubsection*{L515-Indoor} 
\begin{figure}[!t]
	\centering
	\includegraphics[width=0.99\linewidth, trim={1cm 1.5cm 1cm 0.2cm},clip]{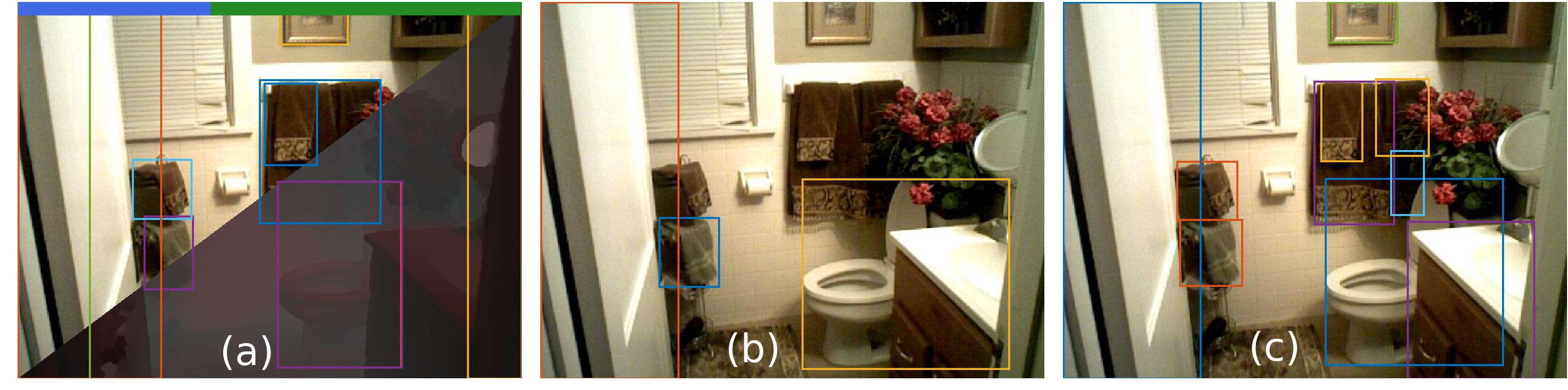}
	\vspace{-0.5cm}
	\caption{(a) Sample output of GEM ($g_\text{sc}$) on the SUNRGB-D dataset with RGB images and ``r-blended'' depth modality. In (b), the output of single modal object detector trained only on RGB images is shown, while (c) is the ground truth. }
	\label{Fig:SUNRGBD-rblended}
	\vspace{-0.6cm}
\end{figure}

Table~\ref{tab:l515} presents the evaluation results of L515-Indoor dataset.
We tested the performance RGB-only and IR-only networks, as well as the $g_\text{sa}$ variant of GEM on this dataset.
Evidently, $g_\text{sa}$ outperformed both single modality detectors
providing an additional functionality of switching between the dominant sensors in changing lighting conditions.
The performance of $g_\text{sa}$ with MobileNet v2 backbone is also presented in the table.
The qualitative results on all three datasets are shown in Figures~\ref{Fig:qualitative_FLIR_L515} and \ref{Fig:SUNRGBD-rblended}.

\begin{table}[!t]
\centering
\caption{Performance evaluation on L515-Indoor dataset.}
\vspace{-0.2cm}
\label{tab:l515}
\begin{tabular}{c|c|c|c|l|c|c} 
\hline
\multirow{3}{*}{\begin{tabular}[c]{@{}c@{}}Model\\w/ RSH\end{tabular}} & \multicolumn{5}{c|}{w/o Random Shadows and Highlights}                                                   & w/ RSH                                                                  \\ 
\cline{2-7}
                                                                       & \multicolumn{4}{c|}{AP@IoU=0.5} & \multirow{2}{*}{\begin{tabular}[c]{@{}c@{}}mAP@\\IoU=0.5\end{tabular}} & \multirow{2}{*}{\begin{tabular}[c]{@{}c@{}}mAP@\\IoU=0.5\end{tabular}}  \\ 
\cline{2-5}
                                                                       & Chair & Cycle & Bin   & Laptop  &                                                                        &                                                                         \\ 
\hline\hline
rgb-only                                                               & 0.909 & 0.912 & 0.920 & 0.911   & 0.912                                                                  & 0.769                                                                   \\
ir-only                                                                & 0.141 & 0.557 & 0.012 & 0.690   & 0.386                                                                  & 0.311                                                                   \\ 
\hdashline
$g_{\text{sa}}$ m-net                                                       & 0.851 & 0.895 & 0.705 & 0.740   & 0.811                                                                  & 0.685                                                                   \\
\hdashline
$g_{\text{sa}}$                                                        & 0.968 & 0.998 & 0.997 & 0.979   & 0.982                                                                  & 0.945                                                                   \\
\hline
\end{tabular}
\vspace{-0.6cm}
\end{table}

\subsubsection*{MobileNet v2} On a mobile platform having a 12-Core Intel\textsuperscript{\textregistered} Core\textsuperscript{TM} i7-9750H CPU, and Nvidia GeForce RTX 2080 GPU, with ResNet-50 backbones, it takes approx 106.0 ms for a single forward pass on the proposed multi-modal object detector.
However, with MobileNet v2 backbone feature extractors, the time for a single forward pass reduces to 49.7\,ms obtaining approximately 20.1 fps.
The drop in prediction accuracy of the deep models with the decrease in the number of network parameters for faster detection speed, is a well-known dilemma (e.g., in our case 23 million parameters for ResNet-50 to 3.4 million parameters for MobileNet v2).
A compromise on prediction accuracy should only be made in non-critical cases where human safety is not at stake.
Otherwise, the use of lightweight backbones should be avoided

\vspace{-.1cm}
\section{Conclusion}
In this paper, we propose GEM, a novel sensor-aware multi-modal object detector, with immunity against adverse lighting scenarios.
Among the proposed sensor fusion configurations, the scalar averaging variant of the deterministic weighted fusion outscored the state-of-the-art and other fusion methods.
The mask multipliers may amplify a certain spatial portion of the feature maps, but could also potentially distort the learned features depth-wise.
Concatenation might be useful in cases where the feature spaces of the utilized sensor modalities differ.
Regarding RGB-D data, the proposed ``r-blended'' hybrid depth modality has proven to be a promising and lightweight alternative to the commonly employed HHA-encoded depth images.
However, instead of employing a fixed blend weight $\alpha$, dynamic adaptation driven by ambient light intensity could demonstrate a more realistic use of the proposed hybrid image.
GEM brings along the shortcomings of~\cite{Carion2020a} in multi-modal object detection setting as well, e.g., it struggles to detect small objects and suffers from the computational complexity of the attention layers.
These issues will be addressed in the future work.
\printbibliography 
\end{document}